# Formulation-Level Auto-Tuning for QUBO-Based Machine Learning: A Case Study Across Multiple Quantum-Inspired Annealers

Naoya Mizuki
Graduate School of Informatics,
Nagoya University
Aichi, Japan
mizuki@hpc.itc.nagoya-u.ac.jp

Takahiro Katagiri
Information Technology Center,
Nagoya University
Aichi, Japan
katagiri@cc.nagoya-u.ac.jp

Daichi Mukunoki
Information Technology Center,
Nagoya University
Aichi, Japan
mukunoki@cc.nagoya-u.ac.jp

Tetsuya Hoshino
Information Technology Center,
Nagoya University
Aichi, Japan
hosino@cc.nagoya-u.ac.jp

***Abstract*—This paper presents an Optuna-based formulation-level auto-tuning framework for support vector machines (SVMs) implemented on multiple quantum-inspired annealers. In an annealing-based SVM, continuous dual variables are discretized and converted into a quadratic unconstrained binary optimization (QUBO) model. This transformation introduces three coupled classes of parameters: representation parameters—the encoding base $B$ and bit depth $K$—which determine numerical range, resolution, and QUBO size; the RBF kernel parameter $\gamma$, which determines classifier geometry; and the equality-constraint penalty $\xi$, which controls feasibility and coefficient balance. We formulate their joint selection as a mixed discrete-continuous black-box optimization problem. The framework has two optimization levels: an inner annealer minimizes the generated QUBO, while an outer Optuna loop reconstructs the formulation in every trial and maximizes validation accuracy. The same solver-agnostic procedure is applied to Fixstars Amplify Annealing Engine, Toshiba SQBM+, and Fujitsu Digital Annealer using TPE and Gaussian-process samplers and is compared with conventional grid search. Experiments on linear and nonlinear classification tasks with 0–20% label noise show mean gains over grid search of approximately 0.8 and 2.1 percentage points, respectively. The results demonstrate that formulation quality and backend capability must be evaluated jointly and that task-level feedback can compensate for discretization, penalty imbalance, and backend-dependent approximate optimization.***

## I. INTRODUCTION

Quantum-inspired annealers have emerged as practical optimization platforms for solving large quadratic unconstrained binary optimization (QUBO) problems on conventional hardware. Unlike physical quantum annealers, they do not require cryogenic operation and can provide large, highly connected binary-variable spaces through GPUs or specialized digital circuits. These properties make them attractive for machine-learning algorithms whose training problems can be expressed in quadratic form.

Support vector machines (SVMs) are a representative example. The SVM dual problem is a convex quadratic program in continuous Lagrange multipliers. To solve it with an annealer, the multipliers must be discretized, encoded with binary variables, and incorporated into a QUBO objective together with an equality-constraint penalty. This transformation introduces several hyperparameters that directly affect both the numerical representation and the resulting classifier. In particular, the encoding base $B$ and bit count $K$ determine the range and resolution of the multipliers, the RBF kernel coefficient $\gamma$ determines the geometry of the decision boundary, and the penalty coefficient $\xi$ controls satisfaction of the SVM equality constraint.

The four parameters $B$, $K$, $\gamma$, and $\xi$ interact strongly, but they do not all play the same role. $B$ and $K$ are representation parameters that determine the multiplier alphabet, numerical resolution, QUBO dimension, and coefficient distribution; $\gamma$ is a model parameter that changes every dense kernel coupling; and $\xi$ is a constraint parameter that balances margin maximization against equality-constraint feasibility. Thus, tuning this formulation is fundamentally different from conventional SVM tuning of only $C$ and $\gamma$. Prior annealing-based SVM studies mainly demonstrated feasibility under fixed encodings or limited parameter grids [1], [2], leaving the joint adaptation of representation, model, and constraint parameters across heterogeneous quantum-inspired backends insufficiently investigated.

Evaluating an annealer with one fixed QUBO formulation can confound two effects: the intrinsic search capability of the backend and the quality of the selected representation. A poor encoding or penalty can make a capable solver appear ineffective, whereas backend-specific adaptation can recover a competitive classifier. We therefore propose a solver-agnostic formulation-level auto-tuning framework. The novelty is not a new Bayesian optimizer; rather, the framework treats the complete QUBO formulation as the tunable object and reconstructs it for every trial. The inner level minimizes QUBO energy on the selected annealer, while the outer Optuna loop uses validation accuracy as task-level feedback to adapt $B$, $K$, $\gamma$, and $\xi$.

We instantiate the framework with a tree-structured Parzen estimator and Gaussian-process optimization and compare them with a conventional grid-search baseline. The same outer-loop procedure is reused without backend-specific redesign on Fixstars Amplify Annealing Engine, Toshiba SQBM+, and Fujitsu Digital Annealer through the Fixstars Amplify SDK. Linearly and nonlinearly separable tasks are evaluated at label-noise rates from 0% to 20% in one-percentage-point increments, allowing us to examine accuracy, robustness, and end-to-end tuning cost under a common trial budget.

The contributions of this work are fourfold. First, we formulate joint tuning of QUBO representation parameters ($B$,$K$), the kernel parameter $\gamma$, and the constraint parameter $\xi$ as a mixed discrete-continuous black-box optimization problem. Second, we introduce a two-level framework in which an inner annealer minimizes the QUBO and an outer task-level loop maximizes validation accuracy, thereby optimizing a family of formulations rather than a fixed solver invocation. Third, we realize the same backend-independent tuning procedure on three heterogeneous quantum-inspired annealers without redesigning the mathematical model. Fourth, we provide a controlled evaluation that separates formulation-search quality from backend behavior by comparing adaptive search with conventional grid search under a unified application-level metric, noise protocol, and execution-time analysis.

## II. BACKGROUND

### A. Quantum-Inspired Annealing and QUBO

The Ising model represents an energy function using spins $s_i \in \{-1,+1\}$, pairwise couplings $J_{ij}$, and local fields $h_i$. Its energy is defined in (1).

$$E(s) = -\sum_{i<j} J_{ij}\, s_i\, s_j - \sum_i h_i\, s_i \quad (1)$$

QUBO instead uses binary variables $q_i \in \{0,1\}$ and minimizes the quadratic objective in (2). The two representations are equivalent through $s_i = 2q_i - 1$.

$$E(q) = q^T Q\, q \quad (2)$$

Physical quantum annealers often have sparse native connectivity and require minor embedding, which may consume many physical qubits per logical variable. The quantum-inspired services used here expose effectively fully connected logical problems and automatically handle backend-specific data structures through the Fixstars Amplify SDK. Amplify AE uses GPU-based search and supports up to 262,144 binary variables (131,072 in a fully connected formulation); SQBM+ is a GPU-based simulated bifurcation service with a very large logical variable capacity; and Fujitsu DA uses dedicated digital circuitry for fully connected QUBO problems [4]–[6].

### B. Characteristics of the Evaluated Annealers

Fixstars Amplify AE is a GPU-oriented annealing engine exposed through the Amplify cloud service. Toshiba SQBM+ implements a simulated bifurcation approach in which coupled nonlinear oscillators have evolved toward low-energy binary states. Fujitsu Digital Annealer uses dedicated digital-circuit techniques for rapid exploration of Ising and QUBO energy landscapes. Although all three accept quadratic binary models, their search dynamics, coefficient handling, scheduling, and service-side execution policies differ.

These architectural differences motivate a backend-independent experiment. The client constructs the same symbolic SVM objective and submits the resulting QUBO through one software layer. Backend-specific tuning is restricted to the common timeout value. Thus, observed accuracy differences primarily reflect the interaction between each solver and the encoded optimization landscape rather than manually customized formulations.

The nominal bit capacities of modern quantum-inspired systems are much larger than those of present quantum annealers, and dense connectivity is particularly useful for kernel SVMs because every pair of training samples contributes to quadratic interaction. Nevertheless, nominal capacity does not directly imply useful application scale. Coefficient precision, service latency, memory required for dense matrices, and the number of repeated solves required by hyperparameter optimization can dominate total cost.

### C. Support Vector Machines

For labeled samples $\{(x_i, y_i)\}$ with $y_i \in \{-1, +1\}$, a soft-margin kernel SVM maximizes the dual objective in (3), subject to $0 \le \alpha_i \le C$ and $\Sigma i\ \alpha_i\, y_i = 0$.

$$Max_{(\alpha)} \sum_i \alpha_i - ½ \sum_i \sum_j \alpha_i\, \alpha_j\, y_i\, y_j\, K(x_i, x_j) \quad (3)$$

The radial basis function (RBF) kernel used in this study is given by (4). The regularization parameter $C$ controls tolerance to margin violations, whereas $\gamma$ controls the spatial range of each training point.

$$K(x_i, x_j) = exp\,(-\gamma\, \|x_i - x_j\|^2\,) \quad (4)$$

*A* conventional solver treats $\alpha_i$ as continuous values. Annealing requires binary variables, so each multiplier is represented with $K$ bits in base $B$ as shown in (5).

$$\alpha_i \approx \sum_{k=0}{}^{K-1} B^k\, q_{ik} \quad (5)$$

Substituting (5) into (3) produces a quadratic polynomial. The equality constraint is incorporated as a squared penalty with weight $\xi$. Consequently, the annealer-based formulation has four principal hyperparameters: $B$, $K$, $\gamma$, and $\xi$.

### D. QUBO Formulation of the SVM Dual

Let $q_{\rm ik}$ denote bit $k$ for training sample $i$. Using the encoding in (5), the QUBO minimized by the annealer can be written in the compact form (6).

$$QUBO(q) = ½ \sum_{i,j} y_i\, y_j\, K(x_i, x_j)\ \alpha_i(q)\ \alpha_j(q) - \sum_i \alpha_i\,(q) + \xi\,(\,\sum_i y_i\, \alpha_i\,(q))^2 \quad (6)$$

Here, $\alpha_i(q)$ is defined by (5). The kernel term creates dense couplings proportional to $y_i\, y_j\, K\,(x_i, x_j)\, B^{k+l}$, whereas the last term penalizes violations of the SVM equality constraint.

The base $B$ and bit count $K$ determine the representable multiplier set. Increasing $K$ improves nominal resolution and range but increases the number of binary variables from $N$ to NK and increases the number of pairwise coefficients approximately quadratically. *A* large penalty $\xi$ improves constraint satisfaction but can compress the relative scale of the classification objective, whereas a small $\xi$ may return low-energy states that violate the dual equality constraint. These competing effects make joint tuning necessary.

After annealing, the returned binary state is decoded into multipliers by (5). Predictions are then made using the decision function in (7).

$$f(x) = \sum_i \alpha_i\, y_i\, K(x_i, x) + b \quad (7)$$

The predicted class is sign $(f(x))$. The bias $b$ is estimated from training samples with nonzero decoded multipliers. Because discretization and approximate optimization can produce small violations, numerically stable thresholding is used when selecting effective support vectors.

### E. Annealing Dynamics and Approximate Solutions

Annealing algorithms search for low-energy states rather than solving the quadratic program through exact convex optimization. In a physical quantum annealer, a transverse-field Hamiltonian is gradually replaced by the problem Hamiltonian, and sufficiently slow evolution preserves the ground state under ideal adiabatic conditions. Quantum-inspired machines reproduce selected aspects of this behavior with classical dynamics, parallel updates, or specialized digital search. Their practical objective is therefore to obtain a sufficiently low-energy state within a limited time rather than to certify global optimality.

For classification, exact ground-state recovery is not always necessary. The SVM decision boundary depends on a weighted sum of kernel functions, and many nearby multiplier vectors can induce nearly identical signs of $f(x)$. Two binary states with different QUBO energies may consequently yield the same test accuracy. This many-to-one relation between optimization state and prediction is one reason heuristic annealing can remain useful even when its raw objective value is not provably optimal.

At the same time, the penalty term creates a delicate energy-scale problem. If $\xi$ is too small, the equality constraint may be violated and the inferred bias can become unstable. If $\xi$ is too large, the constraint dominates the dual objective and compresses meaningful differences among candidate classifiers. Coefficient rescaling performed by a backend may further alter this balance. Automated tuning is therefore not merely a convenience; it is part of the effective numerical formulation.

The finite binary representation also imposes quantization. With $K$ bits and base $B$, only a discrete set of $\alpha$ values is available. A coarse representation may eliminate subtle distinctions among support-vector weights, while a large range can waste resolution on values that are rarely selected. The best representation depends on class geometry, kernel width, and label noise, which explains why $B$ and $K$ cannot be fixed confidently across all experimental conditions.

### F. Parameter Sensitivity and Representation Tradeoffs

The kernel parameter $\gamma$ determines the spatial extent of each training sample in the RBF feature space. Very small $\gamma$ values produce an almost linear and highly smooth decision function, which is adequate for the diagonal task but cannot represent the circular boundary accurately. Very large $\gamma$ values localize the influence of individual samples and can fit label noise. The best $\gamma$ therefore shifts toward smoother values as the corruption rate increases, although the exact trend varies among data realizations.

The classical regularization parameter $C$ has a related role: a large value penalizes training errors strongly and encourages the classifier to accommodate mislabeled points, while a small value permits margin violations and favors a smoother boundary. In the annealer formulation, this balance is represented indirectly through the decoded multiplier range and the equality-constraint penalty. Consequently, $B$, $K$, and $\xi$ jointly play a role analogous to both numerical precision and regularization strength.

Base $B$ changes the spacing of representable multipliers. With $B$=2, adjacent binary patterns yield relatively fine increments over a modest range. With $B$=10, the representation can express larger values but becomes coarse, particularly with only two or three bits. The preferable base depends on whether the task benefits more from precise small weights or from a wider range that can emphasize a few support vectors.

Increasing $K$ from two to three doubles neither accuracy nor cost in a simple way. It adds one binary variable for every training sample, increasing the QUBO dimension by 50%, but the number of pairwise interactions grows roughly with the square of that dimension. The added bit can improve resolution, yet it also creates a larger search space and may make a fixed-time annealing run less effective. This tradeoff is precisely the type of interaction that adaptive hyperparameter search can discover.

Penalty $\xi$ controls feasibility, but its optimal value cannot be chosen solely by minimizing the equality residual. Excessively enforcing $\Sigma_i \alpha_i y_i = 0$ can reduce the solver attention available for maximizing the margin objective. A practical model must balance both components. In future work, a multi-objective tuner could optimize validation accuracy together with constraint violation and QUBO energy, revealing whether a single scalar penalty is sufficient.

Because the parameter effects are coupled, one-dimensional sensitivity plots can be misleading. For example, a large $\gamma$ may be acceptable with a small multiplier range but overfit when paired with a large range. Likewise, a higher bit count may help only when $\xi$ is rescaled to preserve coefficient balance. The Proposed framework (TPE) sampler is advantageous because it learns useful conditional combinations rather than assuming that each variable can be optimized independently.

## III. PROPOSED FORMULATION-LEVEL AUTO-TUNING FRAMEWORK

### A. Two-Level Formulation Tuning

We propose a formulation-level auto-tuning framework for SVMs implemented on quantum-inspired annealers. The framework does not modify the internal search algorithm of an annealer. Instead, it automatically redesigns the numerical representation and energy landscape submitted to the solver by jointly tuning $B$, $K$, $\gamma$, and $\xi$. Every outer-loop trial generates a different kernel matrix, binary encoding, QUBO dimension, coefficient distribution, and constraint balance. The selected backend then solves that QUBO, the binary state is decoded into dual variables, and validation accuracy is returned as the task-level objective.

The framework has two coupled levels. At the inner level, a quantum-inspired annealer approximately minimizes the generated QUBO energy. At the outer level, Optuna maximizes validation accuracy and can therefore compensate for discretization error, penalty imbalance, and backend-dependent approximate search. We instantiate this framework with TPE and Gaussian-process samplers, denoted Proposed framework (TPE) and Proposed framework (GP). Conventional grid search is used as the baseline. The contribution lies in the formulation-tuning layer and closed-loop evaluation, not in claiming a new TPE or GP algorithm.

### *B.* Parameter Classes and Search Domains

The tunable vector $\theta = (B, K, \gamma, \xi)$ contains three parameter classes. Representation parameters $B$ and $K$ define the radix and bit depth of each encoded Lagrange multiplier. We use $B \in \{2, 10\}$ and $K \in \{2, 3\}$; these choices control representable range, resolution, QUBO variable count, and pairwise coupling count. The model parameter $\gamma$ controls the locality of the RBF decision boundary and is searched logarithmically over $[10^{-4}, 10^{3}]$. The constraint parameter $\xi$ weights the squared equality constraint $\xi\ (\Sigma_i\ y_i\ \alpha_i)^2$ and is searched logarithmically over $[10^{-2}, 10^{2}]$. *A* small $\xi$ may yield infeasible multipliers, whereas a large $\xi$ can dominate the margin objective. Unlike ordinary classifier tuning, changing $\theta$ changes both statistical behavior and the optimization model submitted to the annealer.

For the classical RBF SVM, the tuner selects $C$ and $\gamma$ over $[10^{-4}, 10^{3}]$ with logarithmic sampling. The classical problem is therefore a two-dimensional model-tuning task, whereas the annealer problem jointly tunes numerical representation, model geometry, and constraint enforcement. This distinction is central to the proposed framework: the outer loop explores a family of QUBO formulations, not merely alternative settings of a fixed classifier.

For Conventional grid search, $B$ and $K$ use the same discrete sets, gamma uses $\{10^{-4}, 10^{-3}, 10^{-2}, 10^{-1}, 1, 10, 100, 1000\}$, and $x_i$ uses $\{0, 10, 100\}$, resulting in 96 combinations. Proposed framework (TPE) and Proposed framework (GP) use 100 trials. Every annealer call uses a 1000-ms timeout. Parameters are tuned independently for each backend, task type, label-noise rate, and data realization; parameters found for one condition are not transferred to another.

Algorithm 1. Formulation-level auto-tuning for an annealing-based SVM.

```
   Input: training data D_train, validation
   data D_val, number of trials N_trials, and
   search space Ω
   Output: best hyperparameter set P_best
 1 Create an Optuna study S with the specified
   sampler
 2 for i ← 1 to N_trials do
 3     P_i ← SuggestParameters(S, Ω)
 4     Q_i ← ConstructQUBO(D_train, P_i)
 5     q_i ← Solve(Q_i)
 6     α_i ← ComputeMultipliers(q_i, B_i, K_i)
 7     Accuracy_i ← Validate(α_i, D_val)
 8     UpdateStudy(S, P_i, Accuracy_i)
 9 end for
10 P_best ← GetBestParameters(S)
11 return P_best
```

### *C.* Optimization Objective and Model Selection

For trial i, the outer-loop objective is the validation accuracy Accuracy_i. In line 3, the selected sampler proposes $P_i = \{B_i, K_i, \gamma_i, \xi_i\}$ from the search space $\Omega$ using the trial history stored in study S. Line 4 reconstructs the complete QUBO, so every trial may change the binary representation, QUBO dimension, kernel coefficients, and constraint balance. The selected quantum-inspired annealer solves $Q_i$ in line 5, and line 6 decodes the returned binary state into the SVM dual multipliers. The resulting classifier is evaluated on $D_{val}$ in line 7, and its accuracy is registered in the study in line 8. After all trials, the parameter set with the highest validation accuracy is returned. In the implementation, solver failures and timeouts receive a noncompetitive objective value; this exception-handling step is omitted from Algorithm 1 for clarity. The test set is not used during tuning.

### *D.* Why Formulation-Level Tuning Is Necessary

The parameter effects are strongly coupled. $\gamma$ changes every dense kernel interaction; $B$ and $K$ change multiplier range, quantization, and QUBO size; and $\xi$ changes the relative scale of the objective and equality constraint. *A* coarse Cartesian grid allocates evaluations uniformly even when useful configurations occupy narrow, conditional regions spanning several orders of magnitude. The adaptive outer loop instead directs expensive annealer calls toward combinations that produce high task-level utility.

The lowest QUBO energy is not necessarily associated with the highest classification accuracy because different approximate multiplier vectors can induce similar decision boundaries, while a low-energy state may still exhibit an unfavorable constraint or quantization pattern. By optimizing validation accuracy outside the annealer, the proposed framework selects the formulation that best supports the downstream learning task. This task-level feedback separates formulation quality from backend capability and enables a fairer comparison of heterogeneous solvers under an equal tuning budget.

### *E.* Backend Independence and Implementation

The kernel matrix is recomputed for every proposed $\gamma$, and the complete QUBO is rebuilt through the Fixstars Amplify SDK. The returned state is decoded as $\alpha_i = \Sigma_{k=0}^{K-1} B^{k} q_{ik}$, after which the SVM decision function is evaluated on validation samples. The outer-loop interface is independent of the backend's internal dynamics: any solver that accepts the generated quadratic binary model can be substituted without redesigning the tuning logic. Machine time is reported separately from end-to-end trial time, which also includes parameter proposal, QUBO construction, API communication, decoding, and validation.

*A* budget of about 100 trials balances search quality and cloud-service cost. It allows adaptive samplers to pass their initial exploration stage while keeping the full experiment feasible across three backends, two task geometries, 21 noise rates, and five data realizations. Equalizing trial counts emphasizes how efficiently each sampler uses expensive remote evaluations.

The broad logarithmic domains intentionally avoid favorable manual tuning and test the central hypothesis of this work: the quality of an annealing-based learner depends on adapting its formulation, and solver capability should not be judged from one arbitrary encoding or penalty setting. Equal trial budgets allow the comparison to reflect how efficiently each search strategy uses costly remote objective evaluations.

## IV. EXPERIMENTAL SETUP

### A. Data Sets and Noise Model

Two synthetic binary-classification tasks are constructed in a two-dimensional square. The linearly separable task uses $y = x$ as the class boundary. The nonlinear task uses a circle centered at (50, 50) with squared radius 1600. For each task, 1,600 uniformly distributed samples are generated and split into 100 training, 1,000 validation, and 500 test samples. Features are normalized to [0, 1] using statistics from the training set. Label noise is introduced only by flipping the labels of a randomly selected fraction of samples. The fraction ranges from 0% to 20% in increments of 1%. Five independent data sets are generated for every condition.

The linear task isolates the ability to recover a simple margin in the presence of discretization and label corruption. The circular task requires an RBF decision surface and therefore places greater importance on $\gamma$ and on the detailed distribution of decoded support-vector weights. Together, the tasks expose both favorable geometry and a more demanding nonlinear geometry without introducing class imbalance or missing values.

Label noise is introduced by randomly flipping the class of a prescribed fraction of examples. Rates from 0% through 20% are evaluated as one-percentage-point increments. *A* separate noisy data set is generated for each rate rather than merely perturbing predictions after training. Consequently, the classifier must learn from corrupted labels, matching the intended robustness setting.

Five independent data realizations are prepared for every combination of task and noise rate. Averaging over these realizations reduces the influence of a particular random point cloud or split. Training data are intentionally limited to 100 samples because a dense NK-variable QUBO grows rapidly; the larger validation set supports stable parameter selection, and the separate 500-point test set estimates generalization.

### B. Platforms and Software

The client environment uses Windows 11 with Ubuntu 22.04.3 under WSL2, an Intel Core i9-13900KF processor, and 32 GB of DDR4 memory. Experiments are implemented in Python 3.10.12 with Fixstars Amplify SDK 1.3.1, scikit-learn 1.3.2, Optuna 4.4.0, NumPy 1.26.2, and pandas 2.1.3. Annealer execution timeout is fixed at 1,000 ms for all backends to prioritize solution quality and enable controlled comparison.

### C. Evaluation Metrics

Classification accuracy is the primary quality metric because the synthetic classes are approximately balanced. Robustness is assessed from the accuracy curve across noise levels rather than from a single corrupted data set. Timing is reported in two forms. Machine execution time is the backend-reported search duration for the annealers and the scikit-learn fit duration for the classical SVM. End-to-end trial time additionally includes Optuna proposal generation, QUBO construction, API communication, decoding, training, and validation inference.

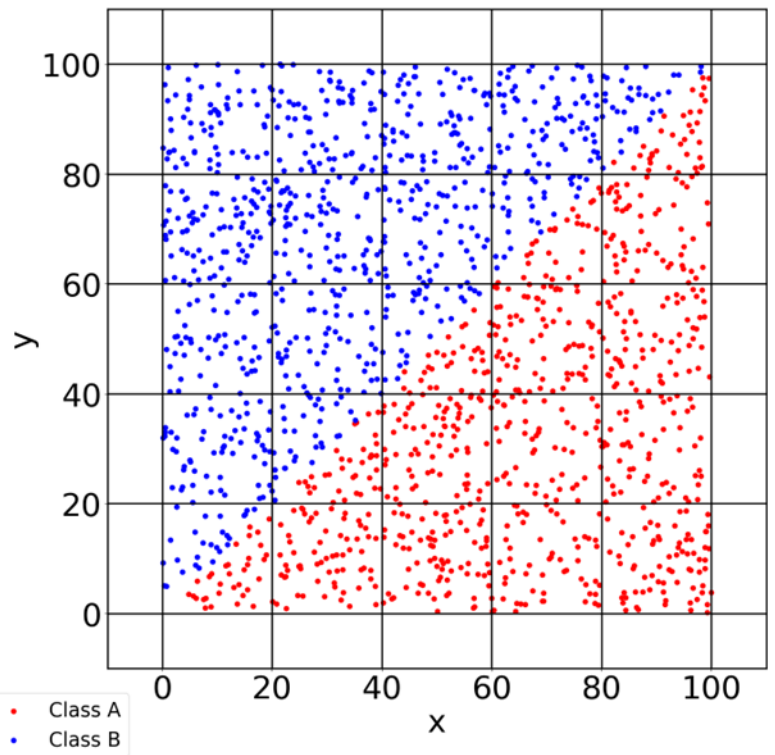

Fig. 1. Linearly separable synthetic data without label noise.

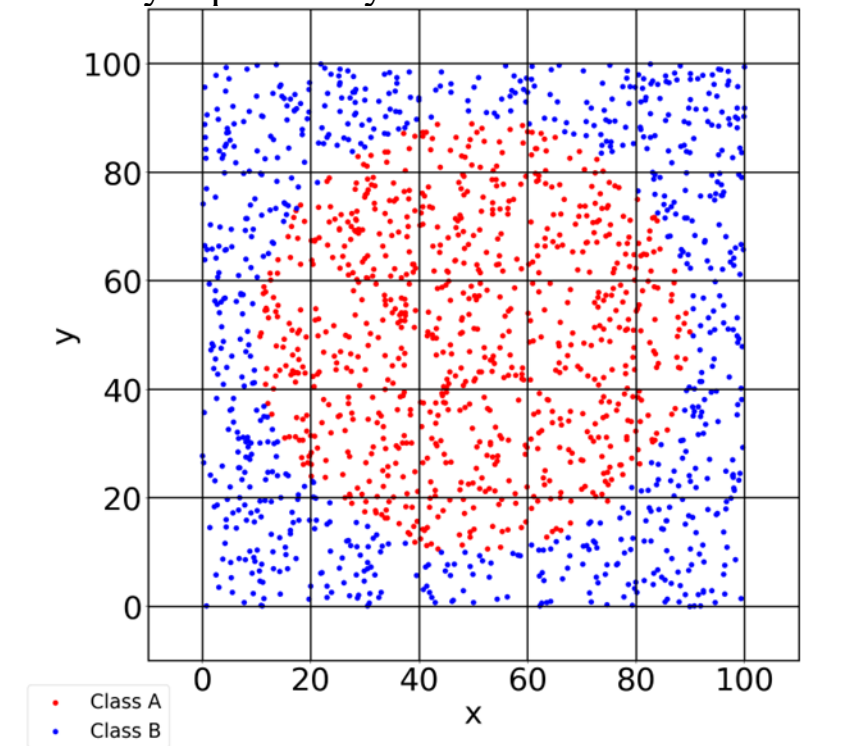

Fig. 2. Nonlinearly separable synthetic data without label noise.

### D. Reproducibility Controls

All methods use identical normalized inputs and the same label realizations. Classical SVM training is performed with scikit-learn using an RBF kernel. Annealer-based training is implemented once through Amplify SDK and switched among providers without changing the QUBO construction. Software versions, client hardware, timeout, trial count, and parameter domains are fixed throughout the study.

### E. Detailed Experimental Protocol

For each task and label-noise rate, a random point set is generated and assigned clean labels from the analytical boundary. The required fraction of labels is then flipped uniformly at random. The data are divided into training, validation, and test subsets before normalization. Minimum and maximum values are computed only from the training subset and applied unchanged to the other subsets, preventing information leakage from validation or test data.

*A* parameter trial trains one classifier on the 100 training samples and evaluates it on the 1000 validation samples. The parameter sets are ranked by validation accuracy. The highest-ranked ten are reconstructed and evaluated on the 500 test samples. Repeating this process for five independent point sets yields a distribution of test accuracies, from which the reported mean curves and aggregate tables are produced.

The unusually large validation set relative to the training set is intentional. Annealer-based training is limited by QUBO

size, whereas validation involves only prediction and is inexpensive. *A* larger validation sample reduces the chance that the optimizer prefers a parameter set because of a few favorable points. The independent test set then protects against repeatedly optimizing the validation metric.

The classical and annealer-based models use the same RBF kernel definition and normalized coordinates. The principal difference is the optimization of the dual variables: a conventional numerical routine solves a continuous constrained problem, while the annealers solve the binary approximation. Therefore, the comparison focuses directly on the consequences of binary encoding and heuristic QUBO optimization rather than on different feature representations.

Every noise level is processed independently. Parameters optimized at one corruption rate are not transferred to another. This avoids giving the optimizer advance knowledge of the trend and reveals whether a method can re-tune itself as the effective data distribution changes. It also produces a demanding but realistic workload in which model selection accompanies each new training set.

Accuracy values are averaged first over selected classifiers and then over the five data realizations. Execution-time measurements exclude one-time environment setup but include operations repeated during normal tuning. Because the services are remote, wall-clock time may include variable network and queue effects; therefore, both machine-reported time and end-to-end time are presented.

# V. RESULTS

## A. Classification Accuracy

Table I summarizes accuracy averaged over all 21 label-noise levels. On the linearly separable task, Amplify AE and SQBM+ with Proposed framework (TPE) both reach 88.0%, while DA reaches 87.6% and the classical SVM reaches 87.1%. The differences among backends are small, indicating that the discretized dual representation can preserve the decision boundary for this simple task. Both proposed methods consistently outperform Conventional grid search.

TABLE I. MEAN ACCURACY ACROSS 0–20% LABEL NOISE: LINEAR TASK (%)

| Backend | Conven-tional (Grid) | Proposal (TPE) | Improvement (point) | Proposal (GP) | Improvement (point) |
|---|---|---|---|---|---|
| Amplify AE | 87.3 | **88.0** | +0.7 | **88.0** | +0.7 |
| Toshiba SQBM+ | 87.1 | **88.0** | +0.9 | 87.9 | +0.8 |
| Fujitsu DA | 87.0 | 87.6 | +0.6 | 87.6 | +0.6 |
| Classical SVM | 86.3 | 87.1 | +0.8 | 87.2 | +0.9 |

For the nonlinear task, the classical SVM is marginally best: 85.3% with Proposed framework (TPE) and 85.6% with Proposed framework (GP). Amplify AE and DA each achieve 85.0% with Proposed framework (TPE) and 85.1% with GP, whereas SQBM+ reaches 84.6%. The gap remains modest, but the result suggests that the four-dimensional annealer hyperparameter space is harder to optimize within 100 trials than the two-dimensional classical space. The nonlinear decision boundary also amplifies small errors in the discretized multipliers.

Tables I and II distinguish the two proposed Bayesian methods from the conventional baseline. The Improvement columns report the absolute percentage-point gain over Conventional grid search for the same backend. Proposed framework (TPE) improves mean accuracy by 0.6–0.9 points on the linear task and 1.4–3.3 points on the nonlinear task. Proposed framework (GP) provides gains of 0.6–0.9 and 1.7–3.3 points, respectively.

TABLE II. MEAN ACCURACY ACROSS 0–20% LABEL NOISE: NONLINEAR TASK (%)

| Backend | Conventional (Grid) | Proposal (TPE) | Improvement (point) | Proposal (GP) | Improvement (point) |
|---|---|---|---|---|---|
| Amplify AE | 83.2 | 85.0 | +1.8 | 85.1 | +1.9 |
| Toshiba SQBM+ | 81.3 | 84.6 | +3.3 | 84.6 | +3.3 |
| Fujitsu DA | 83.2 | 85.0 | +1.8 | 85.1 | +1.9 |
| Classical SVM | 83.9 | **85.3** | +1.4 | **85.6** | +1.7 |

## B. Comparison of Hyperparameter Samplers

Proposed framework (TPE) and Proposed framework (GP) consistently form the strongest method group. On the linear task their mean values are close because many parameter settings recover a nearly correct diagonal boundary. Conventional grid search is slightly weaker, but the absolute differences remain modest. On the nonlinear task, the ordering becomes clearer: the adaptive proposed methods better locate combinations of kernel width, penalty weight, and multiplier encoding that reproduce the circular class boundary.

The results also show that exhaustive enumeration of a coarse grid is not equivalent to effective exploration. Logarithmic ranges spanning several orders of magnitude cannot be represented finely with only a few grid points. Proposed framework (TPE) concentrates trials around favorable scales after observing early outcomes, while retaining sufficient exploration to move between regions. This explains why 100 adaptive trials can outperform 96 nominally exhaustive grid combinations.

Proposed framework (GP) is competitive with Proposed framework (TPE), but its modeling assumptions are less natural for mixed discrete and continuous variables. Proposed framework (TPE) offers a practical balance: it handles conditional or categorical dimensions directly, has low overhead relative to a remote one-second solve, and improves accuracy without requiring problem-specific acquisition-function design.

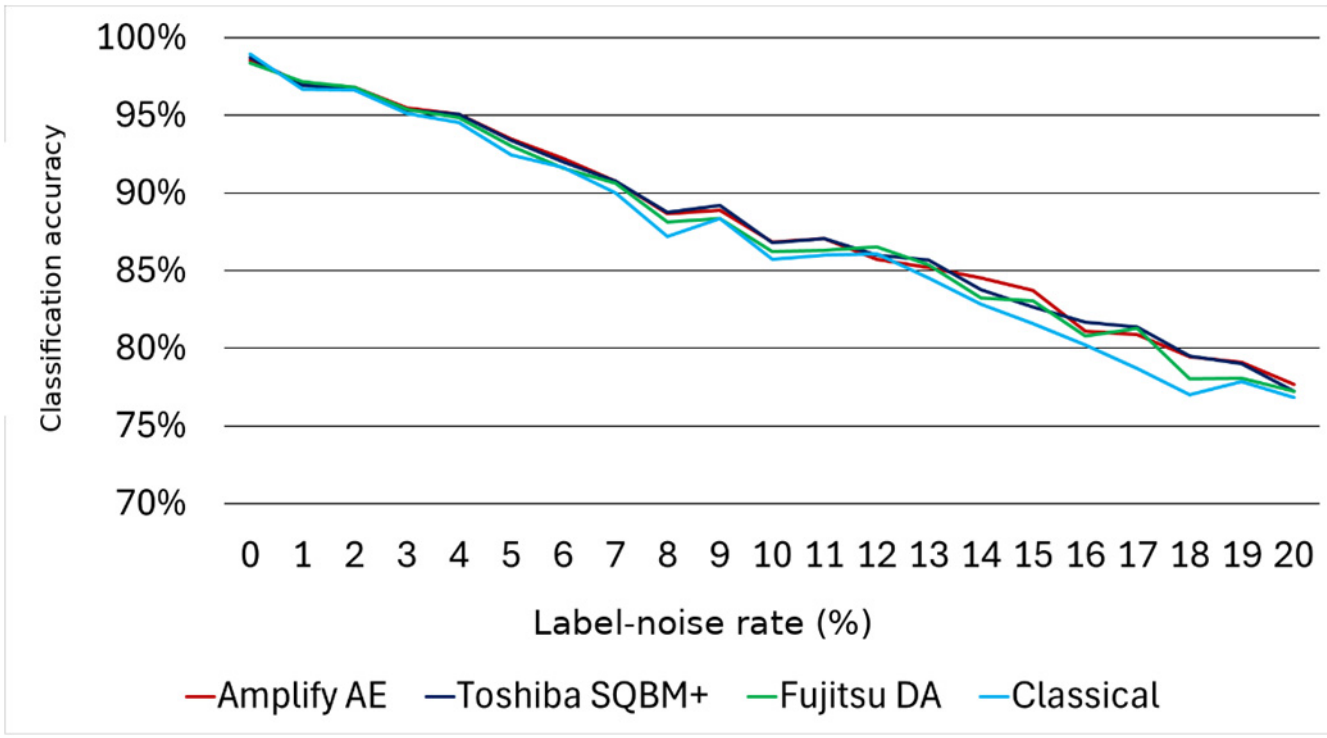

Fig. 3. Accuracy versus label-noise rate for the linearly separable task using Proposed framework (TPE).

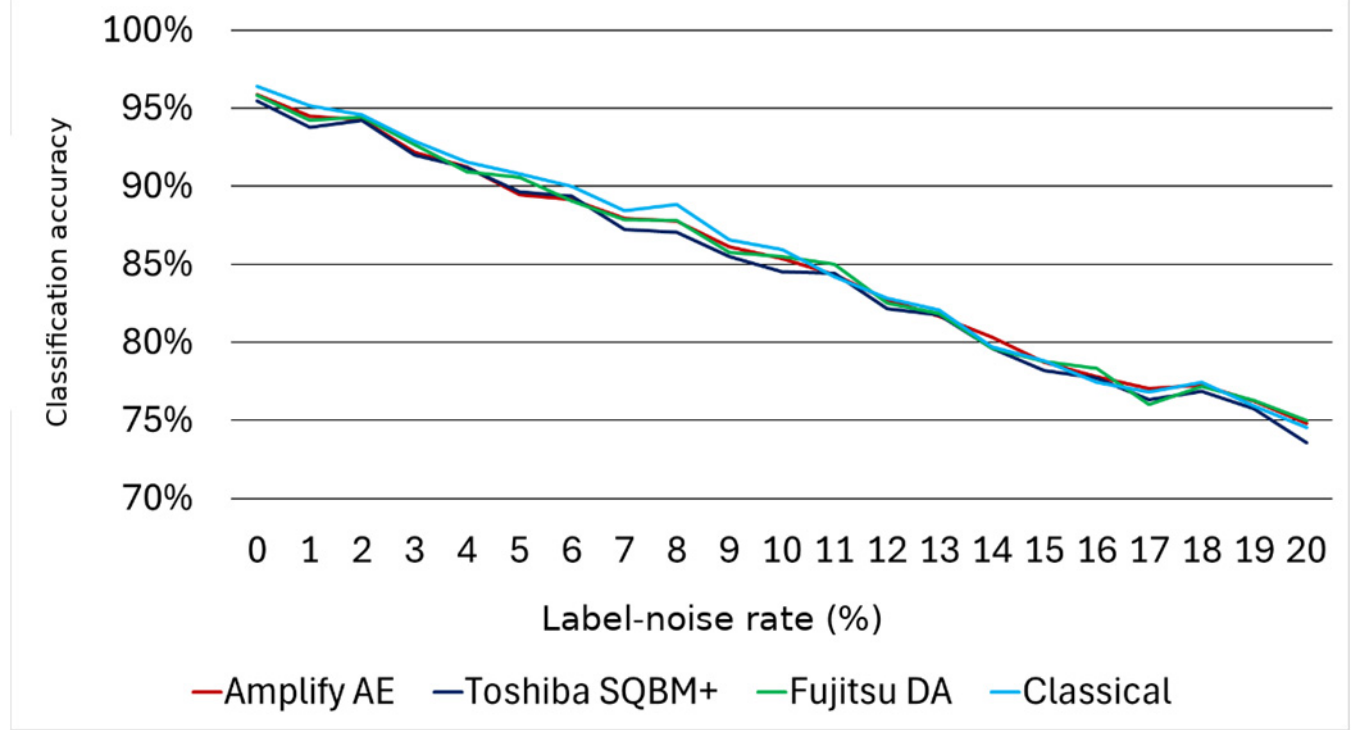

Fig. 4. Accuracy versus label-noise rate for the nonlinear task using Proposed framework (TPE).

### C. Decision Boundaries

Representative decision boundaries obtained with Proposed framework (TPE) are shown in Figs. 5 and 6. For linear data, all three annealers recover a nearly diagonal boundary at zero noise and retain the global trend at 10% and 20% noise. For nonlinear data, the learned boundaries remain approximately circular, although their centers, radii, and local curvature change as corrupted labels increase. These examples confirm that the QUBO solution is not merely fitting a trivial constant classifier.

### D. Robustness to Label Noise

All methods exhibit a gradual, approximately monotonic accuracy decline as the label-noise rate increases. No annealer shows an abrupt collapse at a particular noise level. More importantly, the four curves in Figs. 3 and 4 remain close over most of the range. Thus, under the present soft-margin formulation, annealer-based training provides robustness comparable to conventional SVM training. The behavior indicates that the penalty term and RBF kernel continue to regulate the influence of mislabeled points after binary discretization.

At low noise rates, the maximum-margin principle and RBF smoothing suppress isolated incorrect labels. As corruption increases, mislabeled samples increasingly appear near or across the true boundary and can become support vectors. Accuracy therefore declines for every implementation. The similar slopes indicate that binary multiplier encoding does not introduce a qualitatively different failure mode within the tested range.

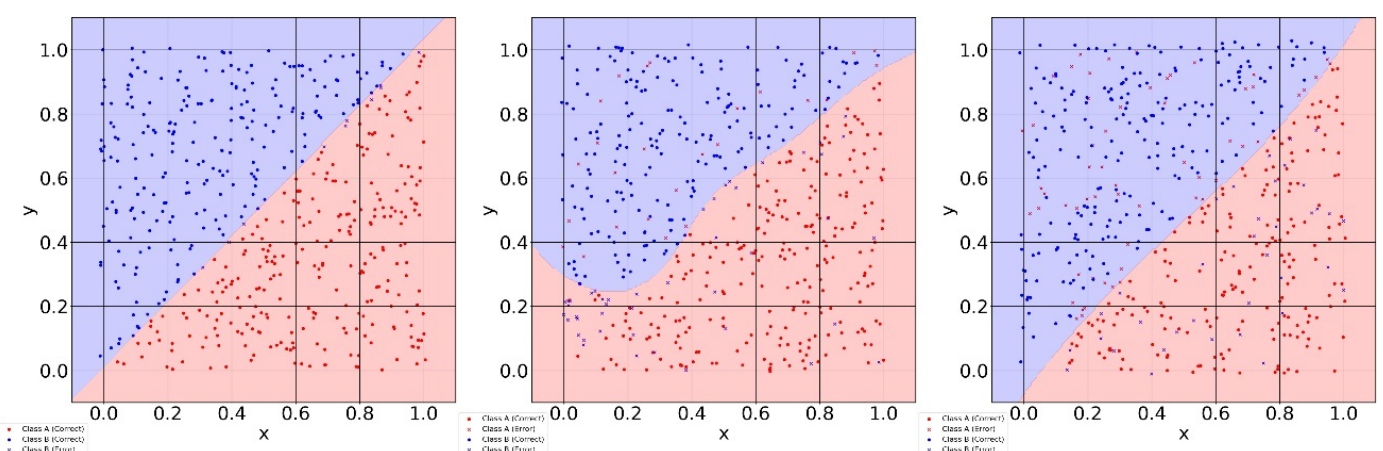

Fig. 5. Amplify AE decision boundaries for the linear task at 0%, 10%, and 20% label noise.

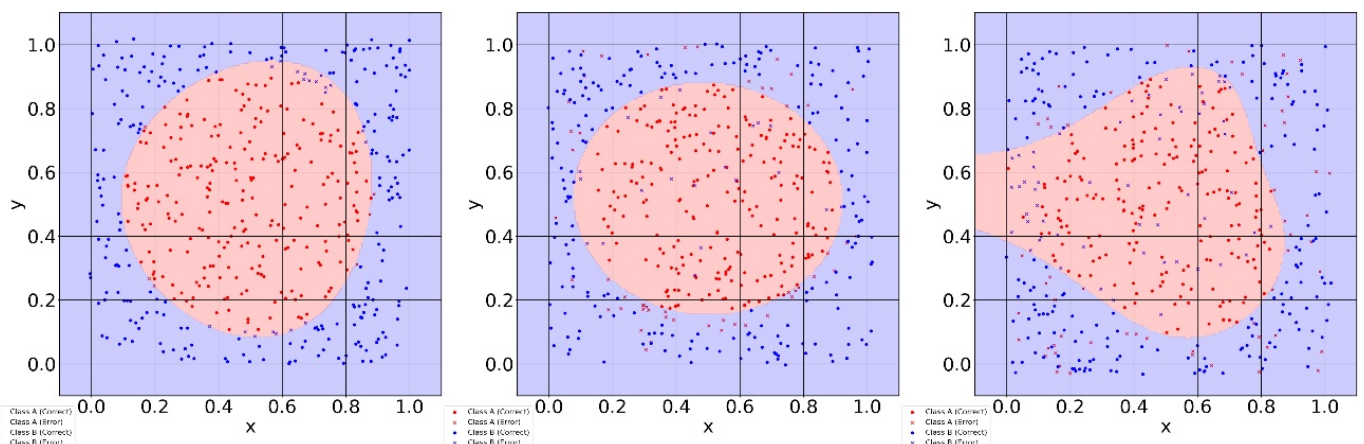

Fig. 6. Amplify AE decision boundaries for the nonlinear task at 0%, 10%, and 20% label noise.

Small crossings among backend curves should not be interpreted as a consistent ranking at individual noise percentages. They are comparable to variation across data realizations and selected parameter sets. The more robust conclusion is that all four environments preserve useful classification performance up to 20% label noise, with the principal differences arising from search quality and nonlinear representation rather than from noise sensitivity itself.

The absolute accuracy decline is expected because labels are flipped before training and evaluation data are generated from the same noisy condition. The relevant observation is the relative slope and separation among backends. Since the curves largely overlap, backend-specific stochastic search does not introduce a substantial additional sensitivity to noise. Hyperparameter search strategy has a more visible effect than annealer selection.

### E. Execution Time

Table III reports machine execution time. The classical SVM requires only 0.62–0.90 ms. Amplify AE takes approximately 920–930 ms, SQBM+ approximately 1,020 ms, and DA approximately 1,250 ms. Under these settings, the classical solver is more than three orders of magnitude faster. This comparison must be interpreted carefully because the annealer timeout was fixed at 1,000 ms and was not minimized for each problem. Nevertheless, even aggressive timeout reduction would have to overcome a very large baseline gap for these two-dimensional, 100-sample tasks.

End-to-end trial time further exposes service overhead. Amplify AE requires roughly 1.48–1.51 *s* per trial, SQBM+ 1.57–1.64 *s*, and DA 3.67–3.74 *s*. The roughly 2.4-*s* gap between DA machine time and complete trial time is mainly attributable to communication, conversion, and service-side overhead. On the classical SVM, most samplers require 5–8 ms per trial, but GP optimization takes about 180 ms because fitting and optimizing the surrogate dominates the negligible SVM training cost.

TABLE III. MACHINE EXECUTION TIME (MS)

| Task/Backend | Conven-tional grid search | Proposed framework (TPE) | Proposed framework (GP) |
|---|---|---|---|
| Linear: Amplify AE | 930 | 920 | 930 |
| Linear: SQBM+ | 1020 | 1020 | 1020 |
| Linear: DA | 1250 | 1250 | 1250 |
| **Linear: Classical** | **0.62** | **0.65** | **0.80** |
| Nonlinear: Amplify AE | 930 | 920 | 930 |
| Nonlinear: SQBM+ | 1020 | 1020 | 1020 |
| Nonlinear: DA | 1250 | 1250 | 1250 |
| **Nonlinear: Classical** | **0.64** | **0.70** | **0.90** |

*A* complete tuning run multiplies these per-trial costs by approximately 100 and by the number of data sets and noise conditions. Thus, even a one-second backend call becomes substantial in a systematic experiment. In contrast, the classical baseline completes individual fits in less than one millisecond for the present 100-sample problem. The measured regime therefore favors the conventional solver by several orders of magnitude in wall-clock time.

### *F. Detailed Accuracy Trends*

On the linearly separable task, zero-noise accuracy approaches the upper nineties for every backend. As noise increases, the curves remain tightly grouped and decline toward the high seventies at 20% corruption. Amplify AE and SQBM+ obtain the highest all-range mean with Proposed framework (TPE), but the difference from DA and the classical SVM is less than one percentage point. In practical terms, all four methods recover essentially the same geometric boundary.

The nonlinear task begins at a slightly lower accuracy because the finite training set must estimate a circular boundary using an RBF expansion. The curves again decline smoothly with noise, reaching the mid-seventies near 20%. Here the classical model has a small advantage in the aggregate table. The difference is consistent with the greater sensitivity of the curved boundary to the precision and constraint quality of the decoded multipliers.

Proposed framework (TPE) and Proposed framework (GP) are usually close to one another, but Proposed framework (TPE) is more consistently strong across both problem classes and all annealers. Conventional grid search is reproducible but limited by its coarse discretization. These observations support Proposed framework (TPE) as the default sampler for subsequent annealer-based SVM studies.

The advantage of Bayesian search is not uniform over noise rates. At very low corruption, many parameter settings achieve near-saturated accuracy, so sampler choice has little effect. At intermediate and high corruption, the model must balance margin smoothness against fitting mislabeled points, and the useful parameter region becomes narrower. Adaptive search is therefore more valuable in the difficult part of the curve.

Decision-boundary visualizations reinforce the numerical results. For the linear task, the learned separating region remains approximately diagonal even at 20% noise, although the boundary becomes slightly curved as the RBF kernel adapts to mislabeled samples. For the nonlinear task, the central region remains recognizable as a circle or rounded polygon. Backend differences appear mainly as local deformations rather than a loss of global structure.

The number and placement of effective support vectors provide an intuitive explanation. Label noise introduces contradictory points near both sides of the ideal boundary, increasing the need for local corrections. *A* continuous solver can distribute weight smoothly across many samples. *A* binary solver approximates these weights with a limited alphabet, which may create small angular or radial irregularities. Tuning $\gamma$ and $\xi$ compensates for much of this quantization.

This comparison does not imply that annealing services can never be competitive. The classical cost grows with sample count, feature dimension, kernel evaluation, and repeated model selection, whereas service latency is partly fixed. However, a crossover can only be established with larger instances, local or lower-latency access, and a formulation that avoids dense quadratic growth. The present measurements provide a necessary baseline for such future scaling studies.

TABLE IV. END-TO-END TIME PER OPTUNA TRIAL (MS)

| Task/Backend | Conven-tional1 grid search | Proposed framework (TPE) | Proposed framework (GP) |
|---|---|---|---|
| Linear: Amplify AE | 1490 | 1480 | 1490 |
| Linear: SQBM+ | 1600 | 1570 | 1620 |
| Linear: DA | 3670 | 3670 | 3740 |
| **Linear: Classical** | **4.91** | **7.59** | **179.60** |
| Nonlinear: Amplify AE | 1490 | 1480 | 1490 |
| Nonlinear: SQBM+ | 1590 | 1570 | 1640 |
| Nonlinear: DA | 3670 | 3670 | 3740 |
| **Nonlinear: Classical** | **5.10** | **7.71** | **182.14** |

## VI. DISCUSSION

### *A. Effect of Search-Space Complexity*

The classical RBF SVM optimizes two continuous hyperparameters, $C$ and $\gamma$. The annealer formulation optimizes four parameters, including two representation parameters that affect numerical resolution and QUBO size. With only 100 trials, the probability of sampling a near-optimal region decreases as dimensionality and parameter interaction increase. This explains why the annealer methods match or slightly exceed the classical SVM on a simple linear boundary but trail it on the nonlinear boundary. Increasing the trial budget or using conditional search spaces may reduce this gap.

The benefit of Proposed framework (TPE) over Conventional grid search is particularly important. *A* fixed grid allocates many evaluations to insensitive regions and samples logarithmic continuous ranges coarsely. Proposed framework (TPE) instead concentrates evaluations near parameter combinations associated with high validation accuracy. The observed 2.1-point improvement on the nonlinear task is substantial relative to the small differences among annealer backends. Thus, claims about annealer quality should control for the hyperparameter optimizer; otherwise, poor tuning may be mistaken for poor solver capability.

### *B. Interpretation of Backend Similarity*

Despite different internal algorithms, the three quantum-inspired systems produce closely grouped accuracy curves. One explanation is that the SVM QUBO has a broad basin of near-equivalent low-energy states: several decoded multiplier vectors yield similar decision boundaries even when their objective energies differ. Classification accuracy is therefore less sensitive than exact QUBO optimality.

*A* second explanation is that hyperparameter tuning compensates for backend-specific behavior. For example, a solver that favors a different coefficient scale may still achieve competitive accuracy when Proposed framework (TPE) adjusts $\xi$, $\gamma$, $B$, and $K$. This is desirable from an application perspective, but it also means that comparing solvers at a single fixed parameter setting could be misleading.

### *C. Scaling Limitations*

The present experiment uses only 100 training samples because the dual SVM QUBO introduces $K$ binary variables per sample and dense pairwise couplings. QUBO construction and storage therefore grow rapidly with the number of training points. Preliminary attempts with 5,000 or more samples encountered memory limitations during preprocessing, before the annealer itself became the bottleneck. This is a critical systems issue: a backend may search a large QUBO efficiently, yet the classical pipeline that forms and transfers the dense coefficient matrix can dominate cost.

Feature dimensionality may provide a more favorable regime than sample-count scaling. Previous work suggests that annealing-based SVMs may become relatively attractive as the number of features increases and classical kernel processing becomes expensive [9]. The current two-dimensional problems are intentionally transparent but are exceptionally easy for conventional CPUs. Future evaluation should therefore use tens to hundreds of dimensions, real data sets, controlled class imbalance, and timeout sweeps. Sparse or low-rank QUBO construction, decomposition, and batch submission should also be investigated.

Several formulation improvements could extend scale. Sparse or low-rank kernel approximations can reduce dense interactions; decomposition can solve subsets of variables iteratively; and adaptive bit allocation can assign more precision only to likely support vectors. Warm starts from a classical approximation may also reduce the annealing budget. These strategies should be evaluated together with accuracy because aggressive compression can alter the margin.

### *D. Practical Implications for Quantum-Inspired Machine Learning*

The results suggest that application quality should be evaluated independently from raw optimizer benchmarks. *A* solver that does not return the lowest measured QUBO energy may still produce an equally accurate classifier, while a slightly lower energy can correspond to an unhelpful change in constraint violation. End users should therefore monitor task-level metrics, feasibility, and stability rather than energy alone.

Automated parameter optimization is essential when comparing heterogeneous annealers. Fixing one shared parameter set may unfairly favor the coefficient scaling or dynamics of a particular machine. *A* fair application-level benchmark should either tune each backend separately under an equal budget or explicitly study transferability of parameters. The present work follows the first approach and thereby measures the best behavior obtainable through a common, realistic tuning process.

### *E. Recommendations for Future Benchmarking*

*A* reproducible benchmark should publish the exact QUBO construction, coefficient normalization, decoding rule, timeout semantics, and treatment of infeasible states. Reporting only the solver name and final accuracy is insufficient because small implementation details can materially change the energy landscape. The common SDK used here reduces interface variation, but backend-specific compilation and scaling may still remain.

Benchmarks should distinguish at least three time components: local model construction, service or communication overhead, and machine search time. Combining them into one number can hide the source of a performance bottleneck. For small instances, overhead dominates; for larger dense instances, QUBO construction and transfer may become substantial; and only beyond that point can raw solver throughput determine total performance.

*A* hybrid strategy offers a realistic route to larger data. Classical screening can identify a subset of likely support vectors, reduce redundant samples, or approximate the kernel matrix. The annealer can then optimize the remaining discrete weights or select among candidate models. Such a design uses conventional numerical methods for operations they handle efficiently and reserves annealing for the combinatorial part of the pipeline.

### *F. Threats to Validity*

First, synthetic data provides controlled geometry and noise but do not reproduce the heterogeneity of real applications. Second, a single client system and specific software versions are used. Third, cloud-service behavior may vary with load, network conditions, and backend updates. Fourth, the 1,000-ms timeout favors consistent solution quality rather than minimum latency. Fifth, accuracy is averaged across five random data sets, which reduces but does not eliminate stochastic variation. Finally, the comparison focuses on predictive accuracy and elapsed time; energy consumption and monetary cost are outside the scope of this study.

The test protocol also uses balanced binary classes and symmetric label flips. Real data may contain asymmetric annotation errors, outliers, duplicated samples, or covariate shift. Accuracy alone can hide class-specific degradation, so future studies should report precision, recall, F1 score, calibration, and area under the ROC curve where appropriate.

Statistical uncertainty should be characterized more explicitly in larger studies. Five data realizations reduce random bias, but confidence intervals and paired significance tests would help distinguish persistent backend effects from run-to-run variability. Repeated service calls at different times would also quantify queue and network variance.

# VII. Related Work

## A. Auto-Tuning Frameworks and Empirical Optimization

Software auto-tuning uses measured performance to search implementation, algorithmic, or runtime parameters. Representative systems include ATLAS for empirical kernel selection [15], Active Harmony for online adaptation [16], FIBER for multi-stage tuning [14], ABCLibScript for directive-based numerical-software tuning [17], and OpenTuner for combining multiple search techniques [18]; a broader survey covers Japanese auto-tuning languages and FFT applications [13]. Our study extends this feedback loop to optimization-model construction. Each trial changes QUBO precision, dimension, coefficients, constraint balance, and downstream accuracy. Established Optuna samplers provide the search engine, while the proposed layer reconstructs the complete formulation, executes the annealer, decodes the state, and evaluates task-level quality.

## B. Annealing-Based SVMs and Positioning

Willsch et al. demonstrated binary SVM training on D-Wave [1], and later work evaluated SVMs on quantum-inspired annealers [2]. These studies established feasibility but mainly used fixed encodings or limited grids. Our framework jointly treats representation, kernel, and penalty parameters as a mixed-domain problem, rebuilds the QUBO in every trial, and applies the same validation-driven procedure to three backends. This differs from classical robust SVM methods that modify the loss or sample weights: we retain the standard RBF-SVM objective and examine whether its robustness survives discretization and approximate QUBO optimization.

## C. Broader Applicability

The framework can be reused for other QUBO-based learning tasks and can tune timeout, coefficient scaling, bit allocation, approximation rank, or decomposition size. It complements conventional software auto-tuning: implementation tuning preserves the mathematical problem, whereas formulation-level tuning changes the numerical encoding and energy landscape. The claimed novelty is this solver-agnostic task-level adaptation and cross-backend evaluation methodology; TPE and GP remain established engines used to instantiate it.

# VIII. Conclusion

This paper presented formulation-level auto-tuning for SVMs on three quantum-inspired annealers. The framework does not introduce a new Bayesian optimizer; it treats the complete QUBO representation as tunable and jointly adapts representation parameters, kernel geometry, and constraint enforcement. An inner backend minimizes QUBO energy, while an outer loop maximizes validation accuracy, separating formulation quality from backend capability. The quantum-inspired SVMs matched the classical RBF SVM on the linear task and remained close on the nonlinear task. TPE and GP instantiations consistently outperformed grid search; TPE improved mean accuracy by about 0.8 percentage points on the linear task and 2.1 points on the nonlinear task.

Across 0–20% label noise, annealer-based classifiers degraded smoothly and retained robustness comparable to the classical baseline, although no execution-time advantage was observed for these small problems. Future work will tune solver controls and encoding structure, evaluate high-dimensional real data, and analyze energy quality, constraint violation, uncertainty, and transferability across backends.

# Acknowledgment

This work was supported by JHPCN and HPCI under project jh260017 and by JSPS KAKENHI Grant JP24*K*02945. The authors used OpenAI ChatGPT for English translation, language refinement, and structural editing of the Abstract, Introduction, and Related Work. All technical content, results, interpretations, and final wording were reviewed and approved by the authors.